\documentclass[10pt,twoside]{article}

\usepackage{8OSME-style}
\usepackage{pgfplots}

\title{Amplifying the kinematics of origami mechanisms with spring joints}
\authorlist[Smith]{M. Smith} % the text in square brackets will be shown in the headers; this should be the author surnames
\affiliations{Malcolm Smith \email{smith.malcolmdavid@gmail.com}}

\makeatletter
  \let\runtitle\@title
  \let\runauthor\shortauthor
\makeatother

\begin{document}

\maketitle

\begin{abstract}
  Due to its rigid foldability and predictable kinematics, the reverse fold is the fundamental mechanism behind some of the most well known origami kinematic structures, including the Miura Ori, Yoshimura, and waterbomb patterns. However, the reverse fold only has one parameter to control its behavior: the starting fold angle. In this paper I provide an alternative to the traditional reverse fold—based on the spring-into-action pattern—called the spring joint. This novel rigidly foldable mechanism is able to couple multiple reverse folds into a compact space to amplify the kinematic output of a traditional reverse fold by up to ten times, and to add one parameter for each reverse fold, giving more programmatic control of origami structures. Methods of parameterizing both the starting angle, the path of travel, and the axis of motion are also introduced such that the spring joint can be engineered to any application within compliant mechanisms, deployable structures and robotics. Unfortunately, this versatility comes at the cost of a large buildup of layers, making the spring joint impractical for thick origami mechanisms. To solve this problem, I also introduce a modular alternative to the spring joint that has no additional layers, with the same kinematic properties. Both of these mechanisms are tested as replacements for the reverse fold in both traditional and custom origami structures.
\end{abstract}

\section{Introduction}
\label{sec:Introduction}
\begin{figure}
    \centering
    \includegraphics[width=80mm]{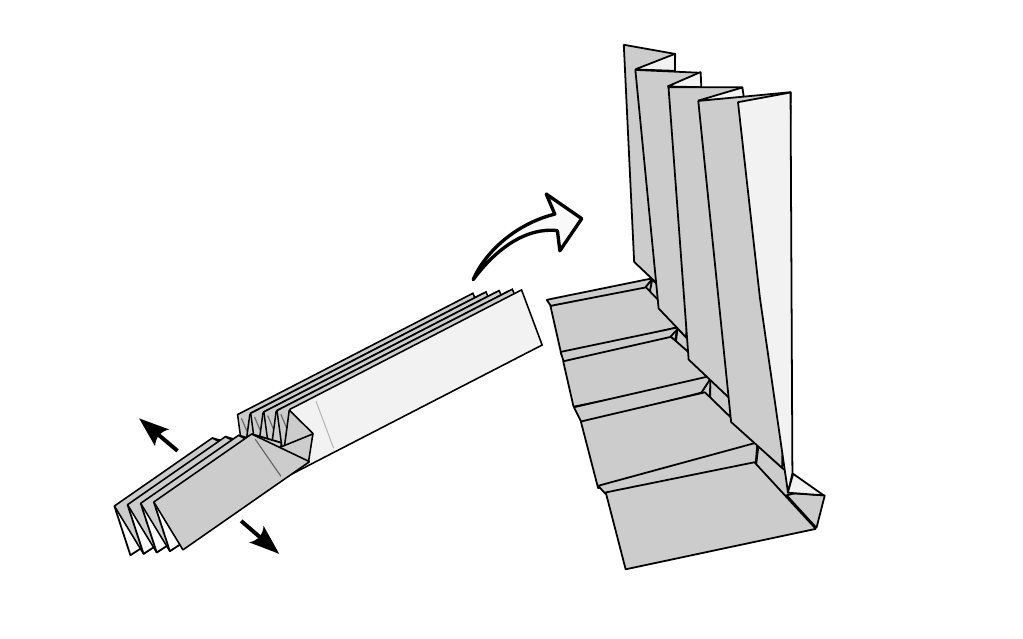}
    \caption{An illustration of the spring joint mechanism when folded from a square of paper}
    \label{fig:motion visual}
\end{figure}

Origami is a vast field of study, from hyper realistic figurative models to fractal tessellations \cite{ikegami2023}. Where origami most closely overlaps with fields of engineering and robotics is in the area of pleat based tessellations known as corrugations. Examples of pleat based origami corrugations that have been used in engineering are the Miura Ori, Yoshimura, Chicken-wire, and Waterbomb patterns. Origami corrugations have proved incredibly useful because they offer a nonrestrictive method of deploying surfaces from compact states to large surfaces. This was the original motivation for the Miura Ori as described in \cite{koryo1985method}. Pleats are also an effective way to transmit mechanical information via the pleat dihedral angle. This is used to facilitate the simultaneous motion of large deployable origami structures. These properties of pleat based origami make it ripe for new innovations and customized designs. However, there has been very little in terms of new applied designs within origami corrugations. Furthermore, the patterns mentioned earlier in this introduction are virtually the \textit{only} corrugation patterns with applications in engineering, are the only templates loaded into the design tool Crane \cite{suto2023crane}, and are–the Miura Ori and Waterbomb fold especially–the only patterns applied to research in aforementioned neighboring fields \cite{lee2021high}\cite{song2014origami}. 

The greatest innovations, then, within pleat based origami have related to the parameterization of existing designs. Tachi \cite{tachi2010freeform} created an algorithm for adapting corrugations to any three dimensional mesh when deployed. Gardiner et al., \cite{gardiner2018fold} developed a design process for adapting in context applications of periodic tessellations. What all of these approaches have in common is that they take existing origami folds and greatly expand their potential for application by adding parameters and program-ability. What all of the ubiquitous designs mentioned have in common is that they are all composed of reverse folds, the most basic unit of corrugations. Reverse folds are simple, but have limited parameters for their rigid motion. The spring joint mechanism (figure \ref{fig:motion visual}) described in this paper offers a parameterized substitution to the reverse fold. The spring joint can amplify the motion of origami mechanisms, and has the parameters such that custom paths of motion can be programmed into the model. Since it is pleat based, it can be integrated to share the dihedral angle motion of any corrugation. Variations and generalizations are introduced such that the spring joint can be applied to real world thick origami applications.  

\section{Construction}
\label{sec:construction}
In this section, I present methods of design and construction based on the intuition behind the kinematics of reverse folds. The term spring joint originates from the Spring-Into-Action model by Jeff Beynon, a model that has been used in origami robotics applications \cite{hu2020origami}. The units described in this paper are a corrugation based generalization of the Spring-Into-Action. The closest to my implementation of this fold, and an inspiration for this paper, is the Squishy Spring-Into-Action Tessellation Tower by Jeremy Shafer\cite{Shafer2013}. The unit that is used in this model can be generalized as a Spring Joint, and the model in this video is similar to the Miura-Ori Pattern with spring joints replacing the reverse folds. 
\subsection{Reverse Fold Kinematics}
\label{subsec:reverse fold kinematics}
For the purposes of this paper, the kinematics of reverse folds will be described with an isotropic generalization, as described in \cite{Tachi2009}. It is isotropic because the degree 4 vertex is parallel along the X axis. Under this framework, the angle of the reverse fold $\phi$ can be modeled as a function of the dihedral fold angle $\xi$, where $\phi_0$ is the initial fold angle as specified by the crease pattern: 
\begin{equation}
    \phi(\xi,\phi_0) = 2arctan\left(cos\frac{\xi}{2}tan\frac{\phi_0}{2}\right)
    \label{eq:single}
\end{equation}
Reverse folds can be constructed in horizontal sequence, simply by connecting two parallel reverse folds with an opposite dihedral angle in between. Any combination of congruent reverse folds will have one degree of freedom, and the kinematic relation continues. The fold angle of the series of reverse folds will have the same kinematic relation between the reverse fold angle and the collective dihedral angles.

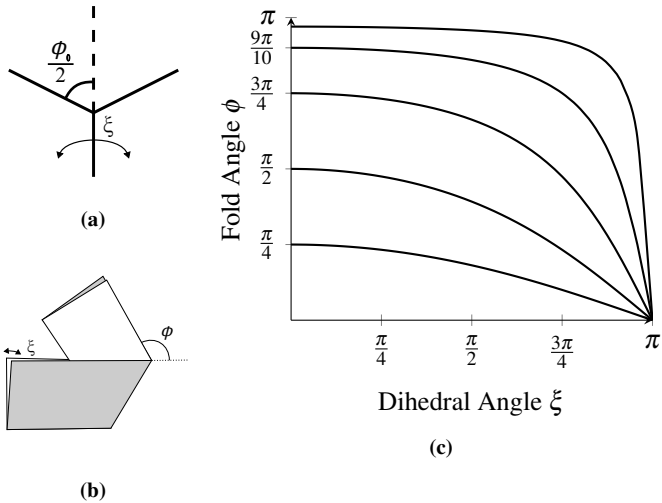
\begin{figure}
\centering
\begin{minipage}{3cm}
  \subfloat[]{
    \includegraphics[width=30mm]{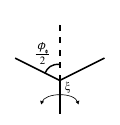} 
    \label{fig:subfig:RFcrease_pattern}
  }  

  \subfloat[]{
    \includegraphics[width=30mm]{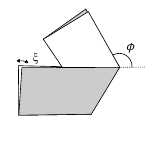}
    \label{fig:subfig:RFfolded}
  }  
\end{minipage}
\begin{minipage}{5cm}
  \subfloat[]{
\begin{tikzpicture}
  \begin{axis} [axis lines=center,
  scale=0.7,
  xticklabels={ $\frac{\pi}{4}$,$\frac{\pi}{2}$,$\frac{3\pi}{4}$, $\pi$},xtick={ pi/4,pi/2,3*pi/4, pi},yticklabels={$\frac{\pi}{4}$,$\frac{\pi}{2}$,$\frac{3\pi}{4}$,$\frac{9\pi}{10}$, $\pi$},ytick={ pi/4,pi/2,3*pi/4,9*pi/10, pi},x label style={at={(axis description cs:0.5,-0.2)},anchor=north},
    y label style={at={(axis description cs:-0.1,.5)},rotate=90,anchor=south},ymax=pi,ylabel={Fold Angle $\phi$},
    xlabel={Dihedral Angle $\xi$},]
  
    \addplot [domain=0:pi, smooth, thick] { 2*rad(atan(cos(deg(x/2))*tan(deg(2.5*pi/20))))};
    \addplot [domain=0:pi, smooth, thick] { 2*rad(atan(cos(deg(x/2))*tan(deg(5*pi/20))))};
    \addplot [domain=0:pi, smooth, thick] { 2*rad(atan(cos(deg(x/2))*tan(deg((7.5*pi)/20))))};
    \addplot [domain=0:pi, smooth, thick] { 2*rad(atan(cos(deg(x/2))*tan(deg((9*pi)/20))))};
    \addplot [domain=0:pi, smooth, thick] { 2*rad(atan(cos(deg(x/2))*tan(deg((9.7*pi)/20))))};
  \end{axis}
\end{tikzpicture}
\label{fig:subfig:RFplot}}

\end{minipage}
\caption{(a) Reverse fold crease pattern (b) Reverse fold folded state (c) Graph of the fold angle $\phi$ as a function of the dihedral angle $\xi$ for different values of the starting fold angle.}
  \label{fig:RF}
\end{figure}

In this model, graphed in figure \ref{fig:RF}, there is a trade off between the travel distance—with higher values of $\phi_0$ traveling further to reach the unfolded state—and the linearity of the relationship between the two variables—with smaller values of $\phi_0$ having a more uniform relationship between $\xi$ and $\phi$. Two things should be noted from this visualization: 1. If the reverse fold is reversed, so that all mountain folds become valleys and vice versa, the kinematics will be reversed, and 2. If you take the limit as $\phi_0$ approaches $\pi$, or the maximum fold angle, the fold angle $\phi$ no longer has any relation to the dihedral angle $\xi$. These folds will be called $\pi$ folds.  

\subsection{Compound Reverse Folds}
\label{subsec:Compound Reverse folds}
The goal of a spring joint is to combine the kinematics of multiple reverse folds. This will amplify the change in the fold angle and increase the design flexibility by increasing the number of variables that can be adjusted. However,  attempts to create a crease pattern that include multiple reverse folds run into the issue that the fold angle of the set of reverse folds will compound, so that the maximum combined fold angle must be less than $\pi$. 
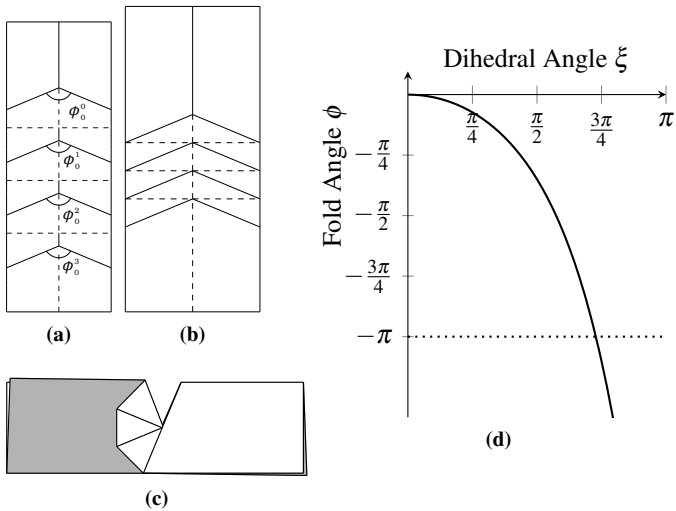
\begin{figure}
\centering
\begin{minipage}{4cm}
  \subfloat[]{
    \includegraphics[width=14mm]{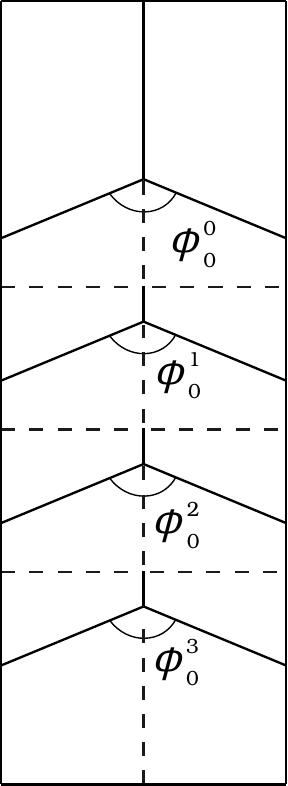} 
    \label{fig:subfig:RFcrease_pattern}
  }  
\subfloat[]{
    \includegraphics[width=18mm]{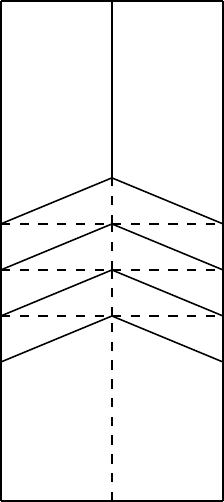} 
    \label{fig:subfig:RFcrease_pattern}
  }  
  
  \subfloat[]{
    \includegraphics[width=40mm]{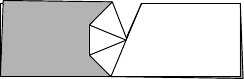}
    \label{fig:subfig:RFfolded}
  }  
\end{minipage}
\begin{minipage}{4cm}
\subfloat[]{
\begin{tikzpicture}
  \begin{axis} [axis lines=center,
  ymin=-4*pi/3,
  ymax=0.3,
  xmax=pi,
  yscale=1.6,
  scale=0.5,
  xticklabels={ $\frac{\pi}{4}$,$\frac{\pi}{2}$,$\frac{3\pi}{4}$, $\pi$},
  xtick={ pi/4,pi/2,3*pi/4, pi},
  yticklabels={$-\frac{\pi}{4}$,$-\frac{\pi}{2}$,$-\frac{3\pi}{4}$, $-\pi$},
  ytick={ -pi/4,-pi/2,-3*pi/4, -pi},
  xticklabel pos=top,
  x label style={at={(axis description cs:0.5,0.25)},anchor=north},
    y label style={at={(axis description cs:-0.2,0)},rotate=90,anchor=south},
    ylabel={Fold Angle $\phi$},
    xlabel={Dihedral Angle $\xi$},]
  
    \addplot [domain=0:pi, smooth, thick] { 4*(2*rad(atan(cos(deg(x/2))*tan(deg(3*pi/8)))))-3*pi};
    \addplot [domain=0:pi, dotted, thick] { -pi};
  \end{axis}
\end{tikzpicture}
}
\end{minipage}
\caption{(a) Compound reverse fold crease pattern with four reverse folds and a starting fold angle of zero. The starting fold angles of the constituent reverse folds $\phi_0$ are shown $\phi_0$. (b) Compressed crease pattern of the compound reverse fold, known as a spring joint. (c) Folded spring joint (d) Graph of the fold angle $\phi$ of this spring joint. A dotted line is drawn to indicate the hard stop of the paper hitting itself}
  \label{fig:SJ}
\end{figure}

A reasonable attempt to solve this issue would be to use reverse folds of the opposite sign to reduce the fold angle. However, if constructed, then the reverse folds of opposite sign will move counter to each other and the actuation is reduced. Instead, we use the reverse folds of the opposite sign with a fold angle of $\pi$ due to their property that the fold angle is unchanging in $\xi$. Using this construction method, each successive reverse fold is paired with a $\pi$ fold of the opposite sign. This process is illustrated in figure \ref{fig:RF}.  In the case of two reverse folds and one $\pi$ fold, the combined fold angle $\phi^{dual}_0$ can be described by: 
\begin{equation}
  \begin{aligned}
    \phi^{dual}_0=\phi_0^0-\pi+\phi_0^1 \\
    \phi^{dual}=\phi(\xi,\phi_0^0)-\pi+\phi(\xi,\phi_0^1) 
  \end{aligned}
  \label{eq:dual}  
\end{equation}
The starting angle of the two individual reverse folds are two parameters that can be used to control the motion of this mechanism. More generally, for any number of reverse folds constructed in this manner:
\begin{equation}
  \begin{aligned}
    \phi^{compound}_0=-\pi (n-1) +\sum^{n}_{k=0}\phi_0^k \\
    \phi^{compound}=-\pi (n-1) +\sum^{n}_{k=0}\phi(\xi,\phi_0^k) 
  \end{aligned}
  \label{eq:compound}  
\end{equation}

\subsection{Parameters}
\label{subsec:Compound Reverse folds}
Figure \ref{fig:SJ}, shows the extent to which motion can be amplified by this process. Much more relative motion can be achieved by combining the kinematics of multiple reverse folds. This method is also very compact, taking up no more space in the final folded model than a reverse fold. For this reason, it can be used as a direct substitution for a reverse fold in applications that require a much higher velocity ratio. Since there is no spacial cost to including more complex spring joints, a reverse fold can be split into multiple separate reverse folds as many times as required for the application. As the limit of the number of reverse folds approaches infinity, and the constituent starting fold angles $\phi_0$ approach $\pi$, the maximum kinematics of a spring joint, given the starting fold angle $\phi_0^{max}$, can be represented as:
\begin{equation}
  \begin{aligned}
    \phi^{max}=(\phi_0^{max}-\pi)sec(\frac{\xi}{2})+\phi_0^{max}\\
  \end{aligned}
  \label{eq:max}  
\end{equation}
The spring joint fold is also extremely programmable. The angle $\phi_0$ of each reverse fold changes the path of the resultant model. For each parameter, then, we are able to add an additional constraint to the path of the motion. For example, we could require that the motion includes two value pairs $(\xi,\phi)$, so long as these values exist in the space below the maximum motion modeled in equation \ref{eq:max} and figure \ref{fig:minimal layer}. We could find a spring joint for an arbitrary starting fold angle $\phi_0$ and a desired fold angle when unfolded $\phi$. The reverse fold function is not sufficiently nonlinear to produce arbitrary functions of $\xi$, but it can still be applied to many applications where a slight "program" embedded into the model itself is beneficial. A larger space of functions is possible when you also control the fold angle of the $\pi$ folds, making the model double back or decouple, but these mechanisms rely heavily on the perfect mathematical model, and do not adequately transfer to the real world. Paper prototypes of decoupling spring joints have only weakly demonstrated this effect, due to the flexibility and memory of the paper, and any compliance in a working model would be compounded in harming this effect. For this reason, the $\pi$ fold angle is not treated as a parameter in the spring joint.

\section{Axis Variations}
\subsection{Tilted Axis Reverse Folds}
Up until this point, I have only discussed reverse folds that follow an axis perpendicular to the pleats, wherein the reverse folds in a sequence are aligned horizontally. Although this mechanism is useful and adaptable, it does not represent the full extent of pleat based rigidly foldable mechanisms. In this section I show that, by using reverse folds at a different axis, functional spring joints can be designed. Increasing the potential designs for spring joints.

\begin{figure}
    \centering
    \includegraphics[width=60mm]{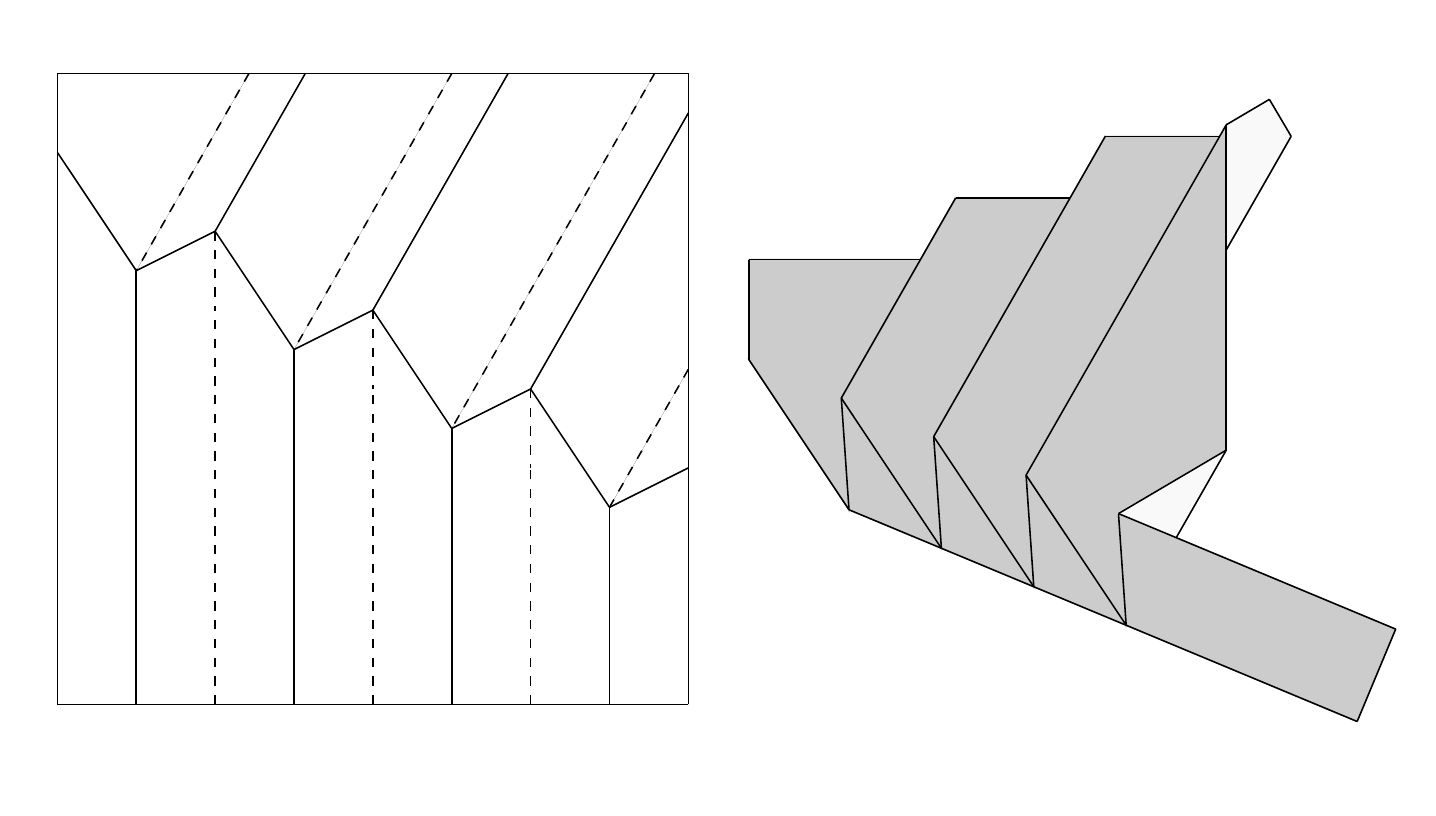}
    \caption{A series of reverse folds, when folded with an axis tilt (and flat folded based on the Kawasaki-Justin theorem), will spread out the pleats on the opposite side of the reverse fold.}
    \label{fig:spread example}
\end{figure}

There is one glaring issue with this model–shown in figure \ref{fig:spread example}: In order to construct the constituent reverse folds of this model, I use the Kawasaki-Justin Theorem \cite{kawasaki1991relation} to generate the fold angles of the resulting pleats. These pleats, however, must spread out under this system and are no longer symmetrical. Any non zero fold angle reverse fold will have this issue. In order for a compound system of reverse folds to preserve the symmetries of the pleats, the initial fold angle of the final system must be zero.
\begin{equation}
    \begin{aligned}
    \phi_0^{tilt}=0
  \end{aligned}
  \label{eq:spread}  
\end{equation}
\subsection{Tilted Axis Spring Joints}
\begin{figure}
    \centering
    \subfloat[]{
    \includegraphics[width=40mm]{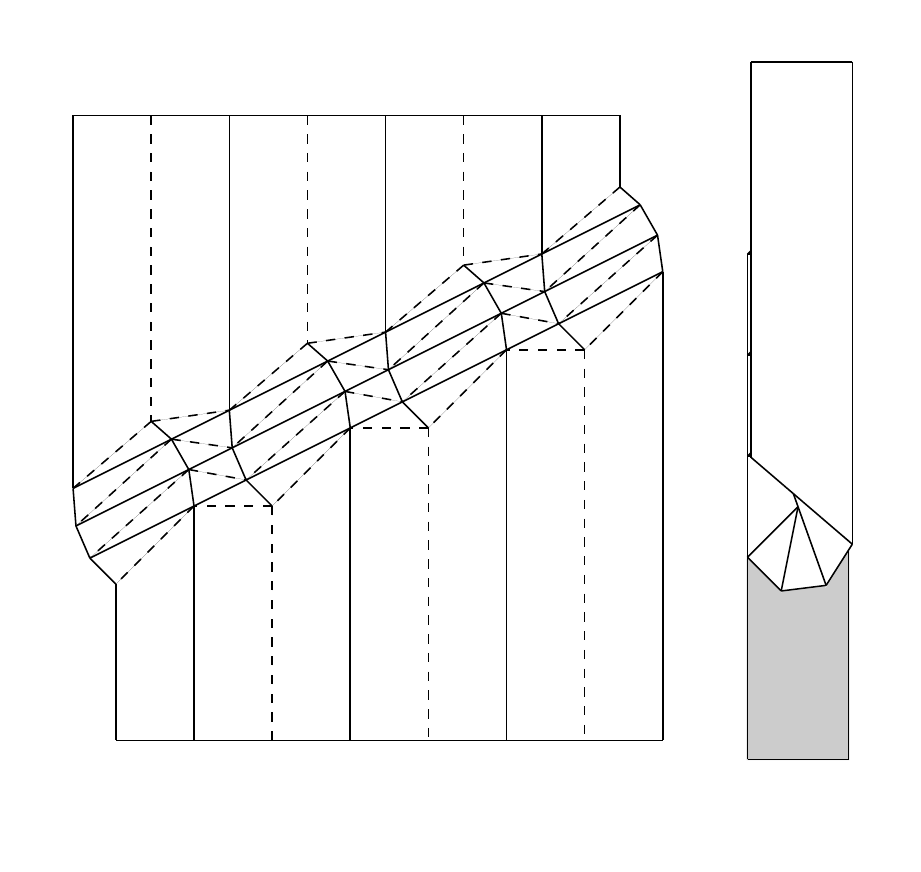}
    }
    \subfloat[]{
    \includegraphics[width=70mm]{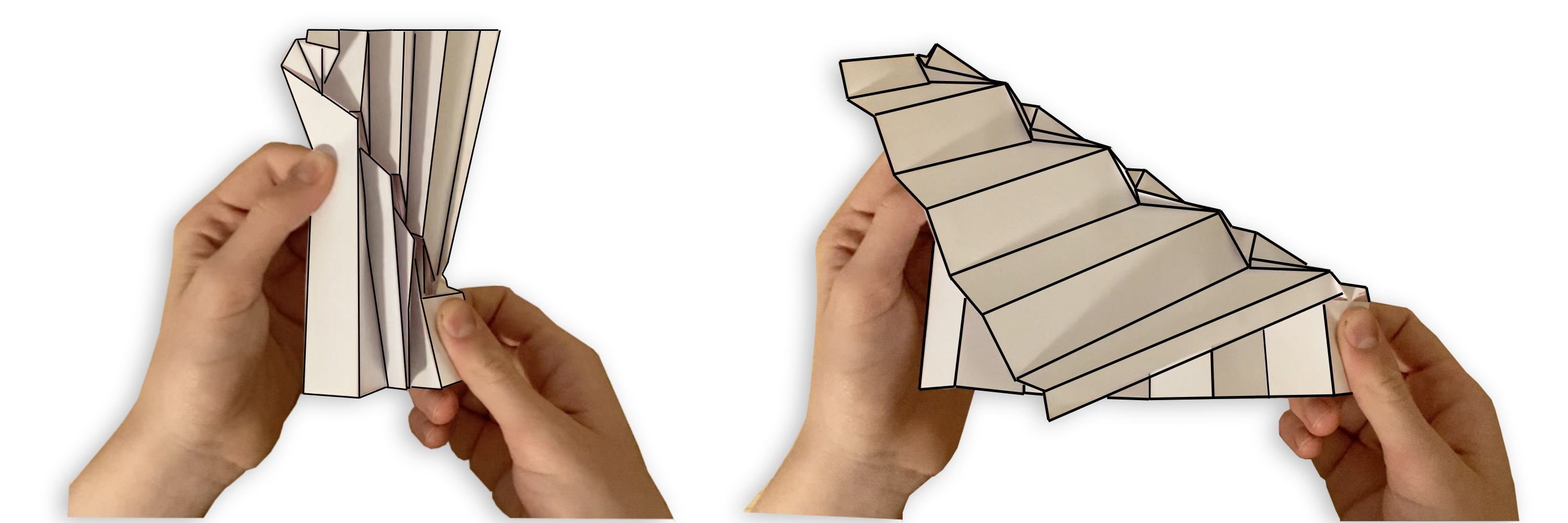}
    }
    \caption{(a) An example tilted spring joint crease pattern and folded state with starting fold angle $\phi^{tilt}_0=0$ (b) Physical prototype where the mechanism actuates along a different axis.}
    \label{fig:tiltSJ}
\end{figure}

By designing crease patterns to hold with this condition, I generated functional spring joints that operate at different angles of rotation. Given that each constituent reverse fold is rigidly foldable, the resultant mechanism is also rigidly foldable. A crease pattern for this mechanism is shown and folded in figure \ref{fig:tiltSJ}. The mechanism, instead of pivoting about an axis perpendicular to the direction of expansion, instead pivots about the tilted axis. When folded flat, the spring joint section of the mechanism is spread out along the paper, rather than be stacked in one area. Because of the complex references and the repeated use of the Kawasaki-Justin Theorem to guarantee flat foldability, there exists no practical method of finding references to fold this model. In my prototypes, I have printed the crease pattern on paper before folding. 

\subsection{Crease Pattern Design}
\begin{figure}
    \centering
    \includegraphics[width=110mm]{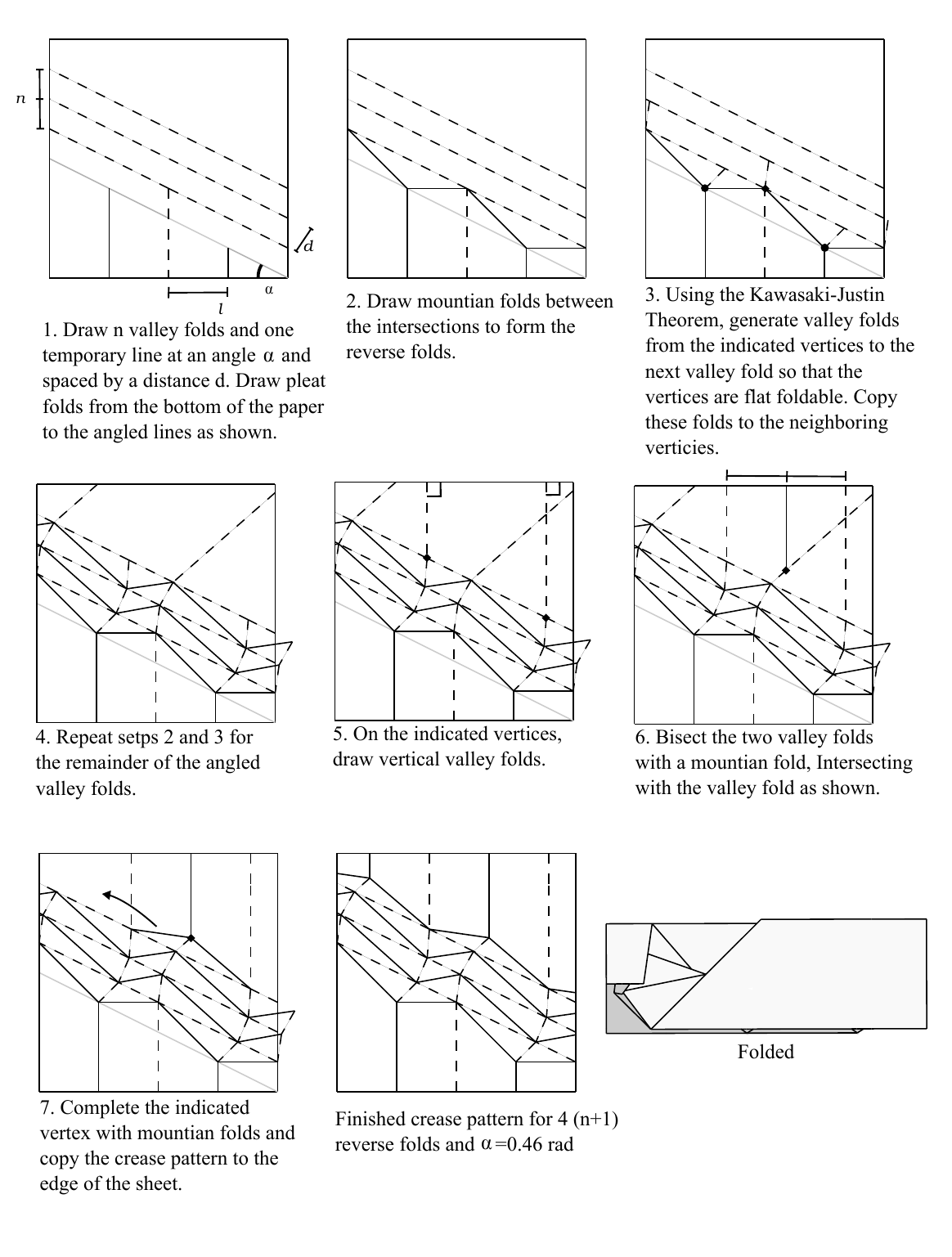}
    \caption{The step by step process for creating a tilted spring joint pattern.}
    \label{fig:tiltDrawSteps}
\end{figure}

The method of generating these crease patterns is detailed in figure \ref{fig:tiltDrawSteps}. The parameters for construction–instead of being the fold angles of the individual reverse folds–are the distances between $\pi$ folds and the number of reverse folds. It is much more difficult to control the fold angle of any one reverse fold, since, the angle of the pleat changes throughout the spring joint area. This can be seen in the curves of pleat lines shown in figures \ref{fig:tiltSJ} and \ref{fig:tiltDrawSteps}. Therefore the fold angle will change depending on the distance between $\pi$ folds and the angle of the pleat, which depends on the fold angle of the reverse folds determined before it, and so on. This iterative process makes it impractical to define individual fold angles by hand. Instead, I define a set distance $l$ between $\pi$ folds and a length $l$ of pleat segments. Generally, the ratio of $\frac{d}{l}$ should be smaller for more reverse folds. Steps 2-4 generate the spring joint pattern based on the new axis and the flat foldability requirement. Steps 5-7 then guarantee that $\phi_0^{tilt}=0$ by forcing the resultant pleats parallel to the initial pleats (5) and bisecting those pleats to avoid spread (6). The crease pattern can be generated based on these restrictions. Computational methods of designing tilted axis spring joints could circumvent this step by solving directly for the final fold angle.
\subsection{Kinematics}
The isotropic assumption used to derive the kinematic model of the basic spring joint requires symmetric around a pleat and so can not longer apply to tilted axis reverse folds. Furthermore, the desired value–"fold angle"–is loosely defined when the mechanism actuates around a different axis, an axis that itself is changing as the pleats open. For these reasons, finding a specific kinematic model to describe the desirable relationships of tilted axis spring joints is an open problem. Nevertheless, the ability to program the axis of rotation into the mechanism serves as another tool to expand the potential applications of this mechanism.
\section{Alternate Modular Design}
\label{sec:modular design}
\subsection{Layer Thickness Accommodation}
Although there is no theoretical limit to the complexity of a spring joint, the multitude of layers makes this model impractical for many real world applications with non zero layer thickness. Of the existing methods for layer thickness accommodation, none of them work with the existing of the spring joint described in figure \ref{fig:SJ}. The tapered panels method described by Tachi in \cite{tachi2011rigid} cannot accommodate full $\pi$ fold angles, and so does not work with the $\pi$ folds of the spring joint. The offset panel technique method by \cite{10.1115/DETC2014-35606} does not extend to work with so many layers on top of each other. And the axis offset method used by \cite{hoberman1988reversibly} and \cite{trautz2010deployable} fails to preserve the kinematics of the original mechanism.
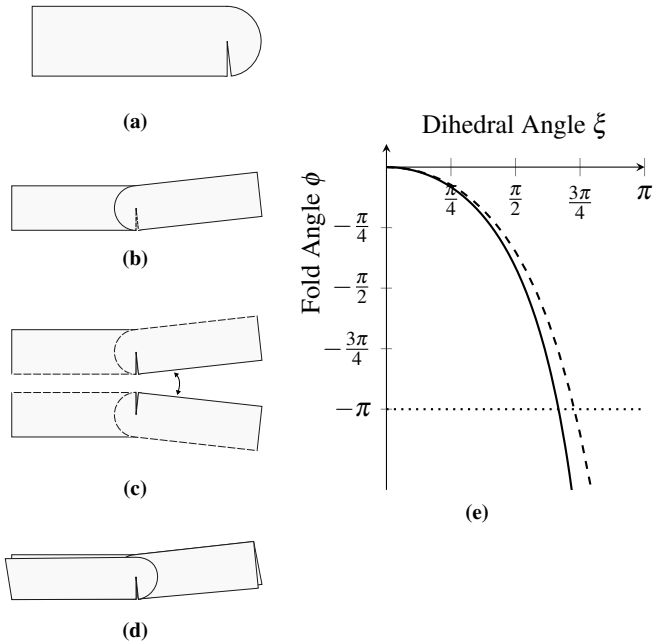
\begin{figure}
    \centering
    \begin{minipage}{4cm}
        \subfloat[]{
        \includegraphics[width=40mm]{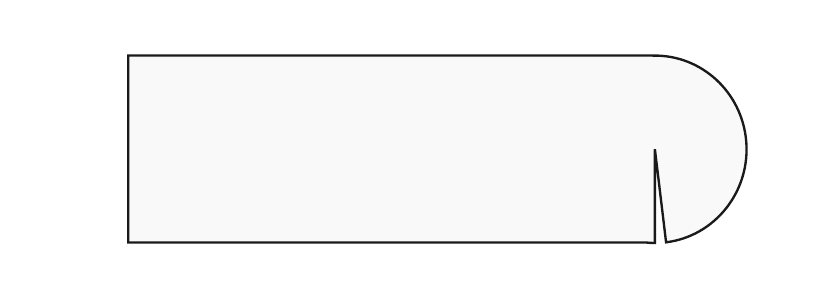}
        }
        
        \subfloat[]{
        \includegraphics[width=40mm]{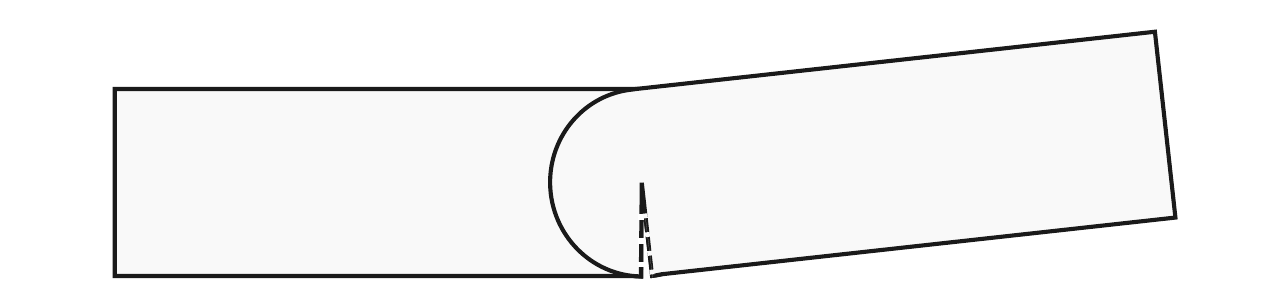}
        }
        
        \subfloat[]{
        \includegraphics[width=40mm]{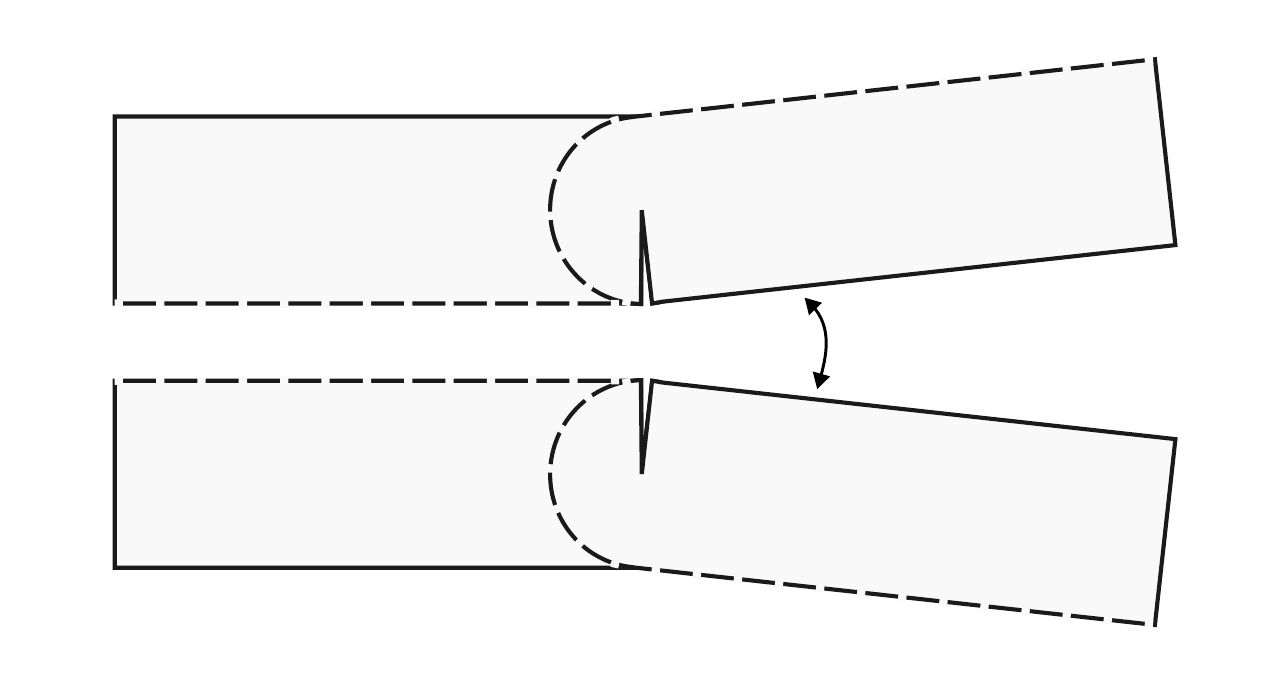}
        }
        
        \subfloat[]{
        \includegraphics[width=40mm]{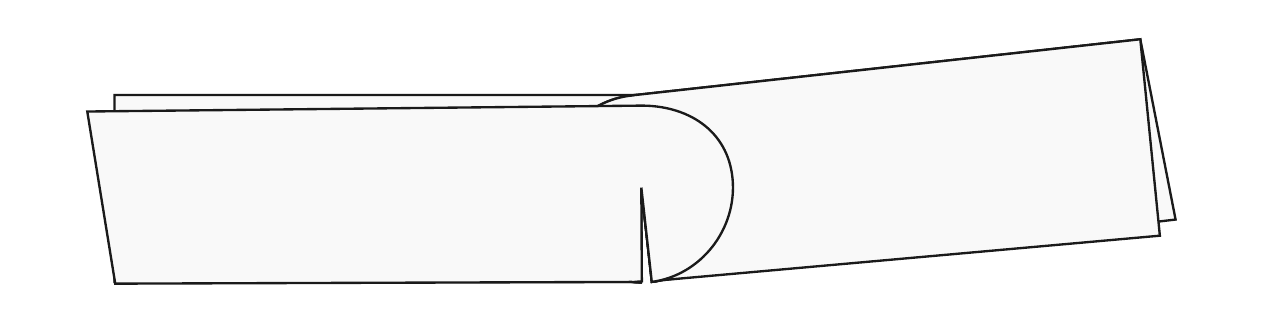}
        }
    \end{minipage}
    \begin{minipage}{6.5cm}
    \subfloat[]{
    \begin{tikzpicture}
  \begin{axis} [axis lines=center,
  ymin=-4*pi/3,
  ymax=0.3,
  xmax=pi,
  yscale=1.6,
  scale=0.5,
  xticklabels={ $\frac{\pi}{4}$,$\frac{\pi}{2}$,$\frac{3\pi}{4}$, $\pi$},
  xtick={ pi/4,pi/2,3*pi/4, pi},
  yticklabels={$-\frac{\pi}{4}$,$-\frac{\pi}{2}$,$-\frac{3\pi}{4}$, $-\pi$},
  ytick={ -pi/4,-pi/2,-3*pi/4, -pi},
  xticklabel pos=top,
  x label style={at={(axis description cs:0.5,0.25)},anchor=north},
    y label style={at={(axis description cs:-0.2,0)},rotate=90,anchor=south},
    ylabel={Fold Angle $\phi$},
    xlabel={Dihedral Angle $\xi$},]
  
    \addplot [domain=0:pi-0.5, smooth, thick] {-pi*sec(deg(x/2))+pi};
    \addplot [domain=0:pi, dashed, thick] { 4*(2*rad(atan(cos(deg(x/2))*tan(deg(3*pi/8)))))-3*pi};
    \addplot [domain=0:pi, dotted, thick] { -pi};
  \end{axis}
\end{tikzpicture}
        }
    \end{minipage}
    \caption{
    Steps for the construction of a minimal layer spring joint: (a) The basic unit for the mechanism (b) Place the two units together as shown, and connect them with hinges along the dashed lines. (c) Connect two mirrored compound units along the dashed line. (d) Finished minimal layer spring joint. (e) The graph of the fold angle (solid) of the minimal layer spring joint compared to the 4 reverse fold spring joint (dashed) from figure \ref{fig:SJ}.
    }
    \label{fig:minimal layer}
\end{figure}
\begin{figure}
    \centering
    \includegraphics[width=60mm]{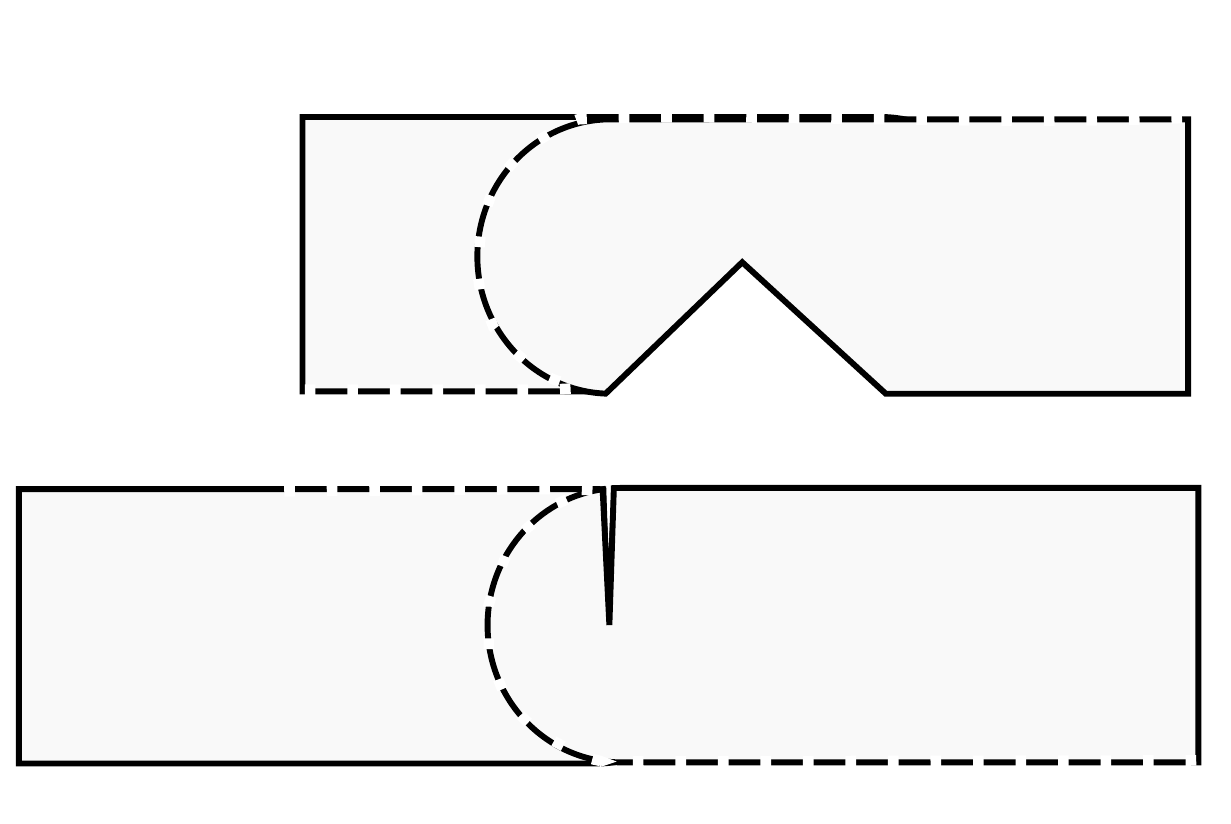}
    \caption{A variation of the modular units and assembly to replicate a tilted axis spring joint. Half of the units have a open cut in replacement of the slit such that, when placed on itself, a greater horizontal displacement is asymmetrically created in one pair of units, causing an axis shift.}
    \label{fig:axis tilt variant}
\end{figure}
\subsection{Minimal Layer Spring Joints}
In this section I introduce an adaptation of the spring joint fold to reduce the layers of the mechanism. The trade off is that this adaptation is not foldable from a planar sheet of paper. Instead, it is constructed by a series of rectangular units with semicircular ends.

A method of construction for this variation is outlined in figure \ref{fig:minimal layer}. In order for this mechanism to operate well, all connections and the semicircular end of the unit must be made with a flexible material. For paper prototypes, scotch tape is sufficient.

This mechanism is also rigidly foldable, and, by replacing the semicircular end with a polygon, it can be made to replicate the motion of any spring joint. Each vertex of the polygon is equivalent a reverse fold-$\pi$ fold pair, where: $\theta^{vertex}=\phi_0$. Therefore–in it's theoretically circular form with infinite vertices–it approximates the limit of spring joint kinematics as described in equation \ref{eq:max}. 

This high velocity ratio is especially beneficial in applications where there is a lack of rigidity in the material–such as origami action models and compliant mechanisms–because it becomes much more responsive and motion is not dampened in the mechanism. The downside to this high velocity ratio, is, of course, a much lower mechanical advantage. For this reason it is not ideal for high force transfer applications, and works well with light materials. The pleats or corrugations also serve to reinforce the material, giving it more structural integrity.
\subsection{Tilted Axis Modular}
Modifications to the original unit, shown in figure \ref{fig:axis tilt variant}, allow for a modular replication of the tilted axis. Unlike the basic minimal layer spring joint, this is not a one to one replica of its origami alternative, but it replicates a similar motion.

\section{Reverse Fold Substitution}
\label{sec:RFsub}
\begin{figure}
    \centering
    \subfloat[]{
    \includegraphics[width=40mm]{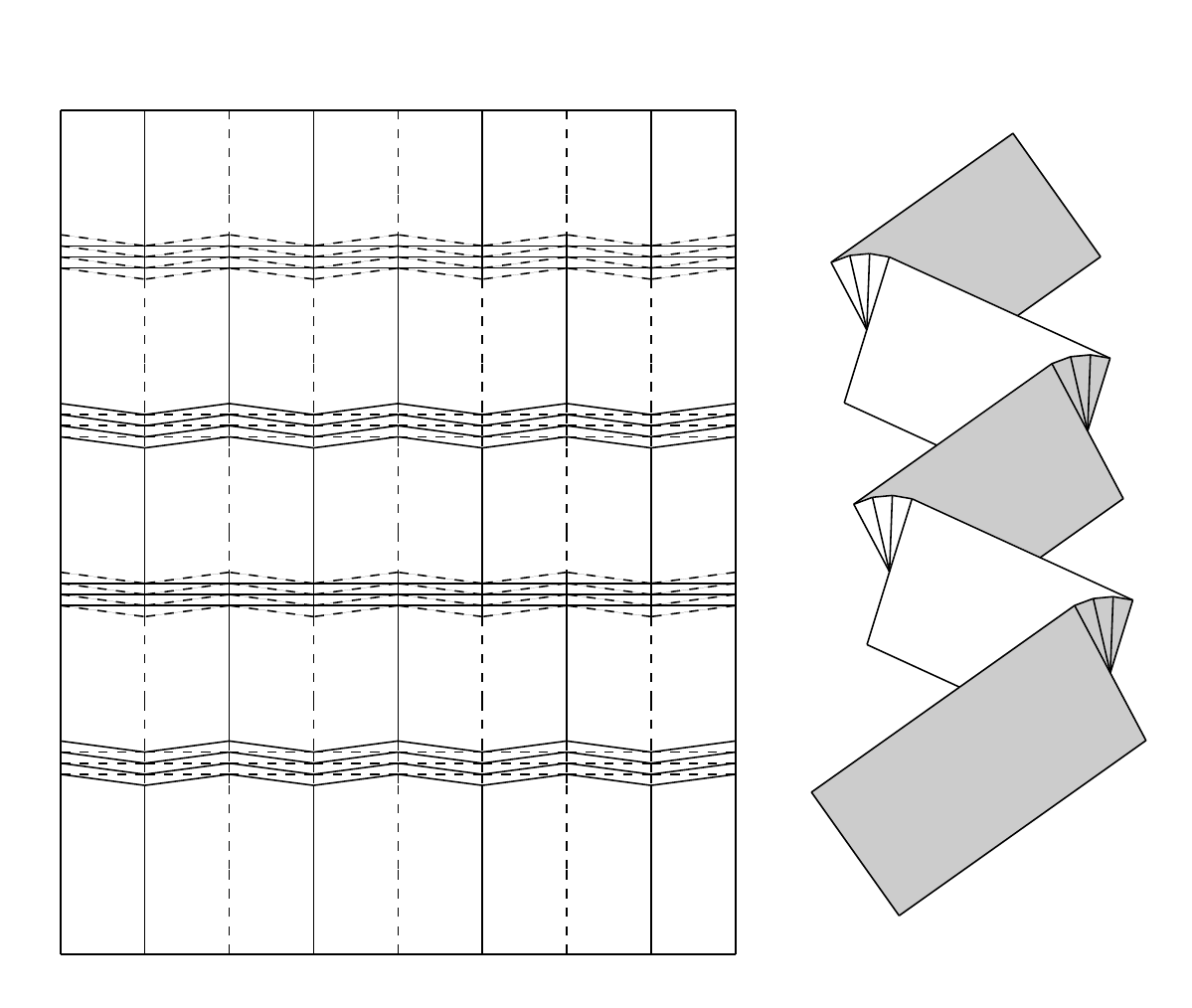}
    }
    \subfloat[]{
    \includegraphics[width=70mm]{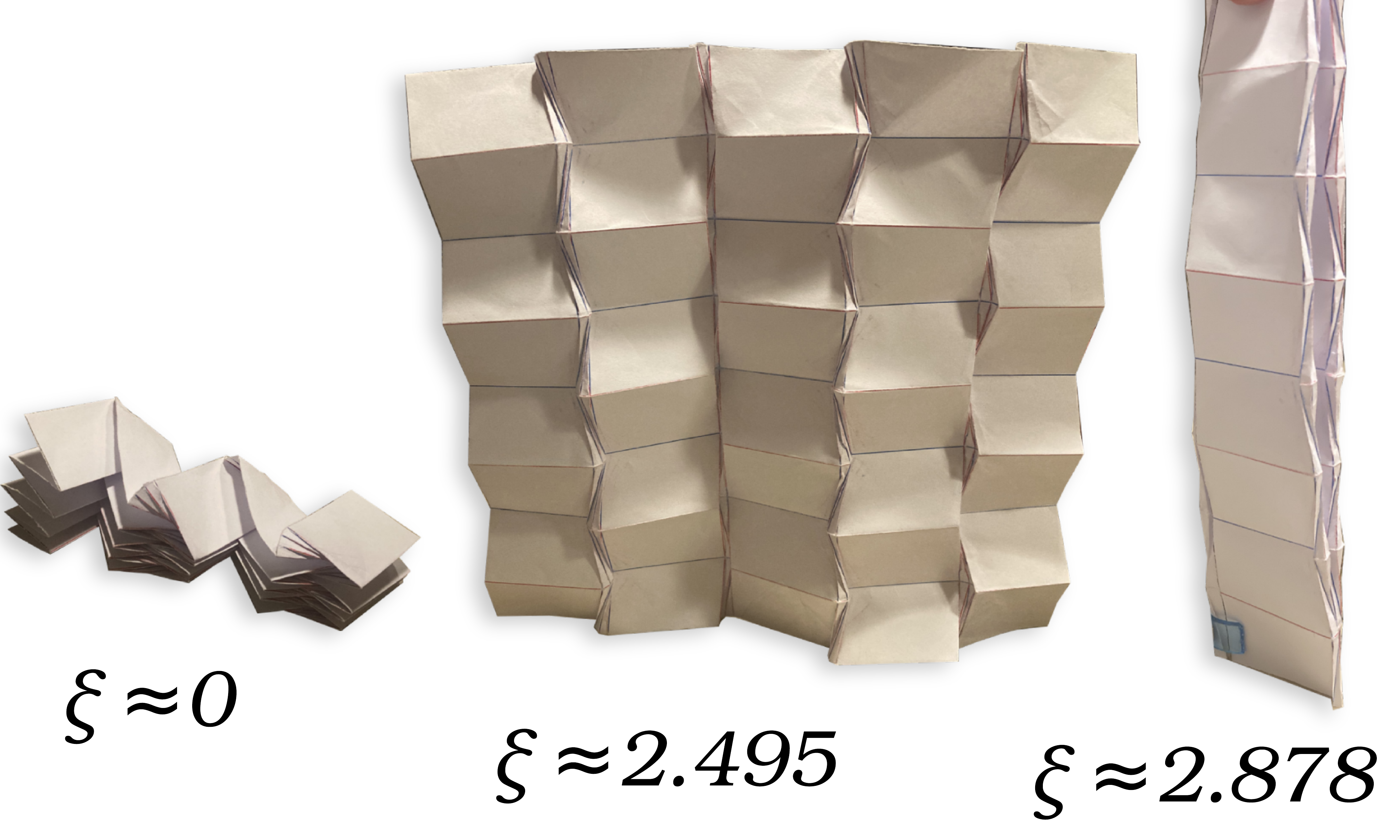}
    }
    \caption{(a) The crease pattern and folded illustration of a Miura-Ori pattern where the reverse folds are substituted by spring joints with $\phi_0^{compound}=\frac{5\pi}{6}$. (b) The physical prototype of this model, shown at varying dihedral angles $\xi$.}
    \label{fig:Miura}
\end{figure}
In origami corrugation design, you can interpret a reverse fold as a segment of the sheet that reverses the pleat sign. Any fold that accomplishes this can be used as a direct substitution for a reverse fold in the crease pattern, with no interference. Since a spring joint is a composition of reverse folds, it has this property and can be used as a substitution for reverse folds to either amplify the motion of a corrugation mechanism, or to display additional properties. 

To illustrate this, I created a paper prototype for a Miura Ori where the reverse folds are substituted for spring joints with $\phi_0^{compound}=\frac{5\pi}{6}$ and four reverse folds each, shown in figure \ref{fig:Miura}. The Miura Ori, first shown in \cite{koryo1985method}, is by far the most popular and widely used origami mechanism. It has the property of a negative Poisson's Ratio, where the model can expand in two orthogonal directions at the same time, making it ideal for the deployment of large surfaces–such as that of solar panels in space. 

My design of the spring joint Miura Ori mechanism had a number of unique properties. Where the fold angle $\phi>0$, the Miura-Ori is deploying as normal, with a negative Poisson's Ratio, in this case when the dihedral angle $\xi<2.495$. At $\xi=2.495$, $\phi=0$, therefore the mechanism is in it's maximum deployed state. In the rest of the domain where $2.495<\xi>2.878$, the model starts to collapse again as $\phi<0$ with a \textit{positive} Poisson's Ratio. When $\xi=2.878$, $\phi=-\pi$, so the pleats collide with each other. These states are shown in figure \ref{fig:Miura}. In this prototype, it is much easier to deploy to the maximum state at  $\xi=2.495$, unlike with paper renditions of the original Miura Ori, where the paper's memory holds the folds in place. This property implies a use in applications of a Miura Ori from a compliant material, like the paper tested here.

\section{Discussion}
\label{sec:disc}
\subsection{Further Development}
Although this paper introduces many variants and generalizations of the spring joint pattern, more generalizations will naturally lead to more applications. The generalizations chosen for this paper were picked for their rigid foldability, and therefore predictable kinematics. Further progress in this field would require exploring spring joint generalizations for non parallel pleats, and a shift in axis in the paper. These generalizations would allow spring joints to adapt to the complex instances of reverse folds seen in freeform tessellation variants \cite{tachi2010freeform}. These generalizations are not rigidly foldable, and cannot be designed by hand. In order to achieve these, computer assisted design tools such as Crane \cite{suto2023crane}, Tree Maker \cite{lang1996computational}, Tess \cite{bateman2002computer} and Origamizer \cite{demaine2017origamizer} are required. In my view, software assisted design and generalization is the surest direction of progress for the spring joint mechanism.

\subsection{In Origami}
\begin{figure}
    \centering
    \subfloat[]{
    \includegraphics[width=40mm]{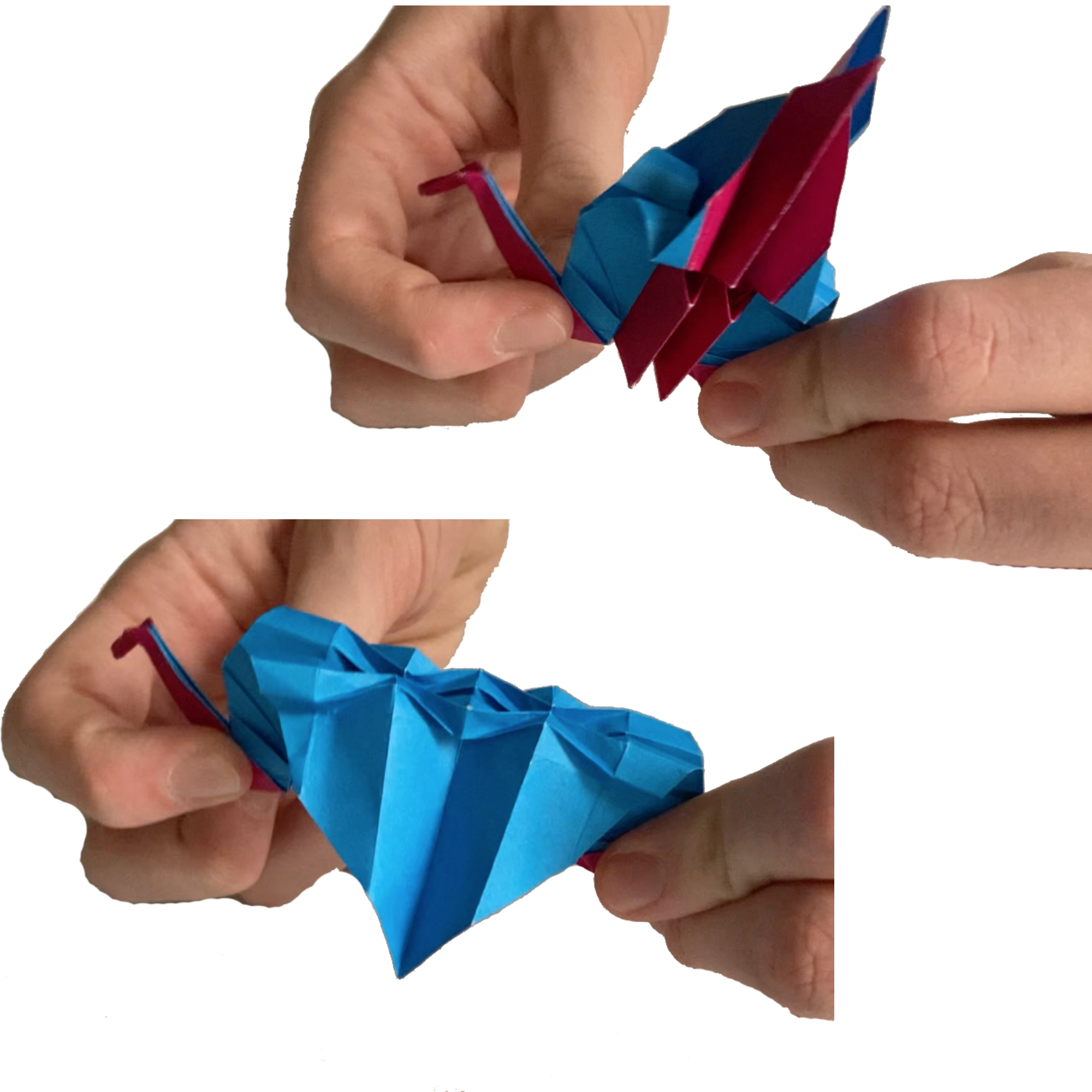}
    }
    \subfloat[]{
    \includegraphics[width=40mm]{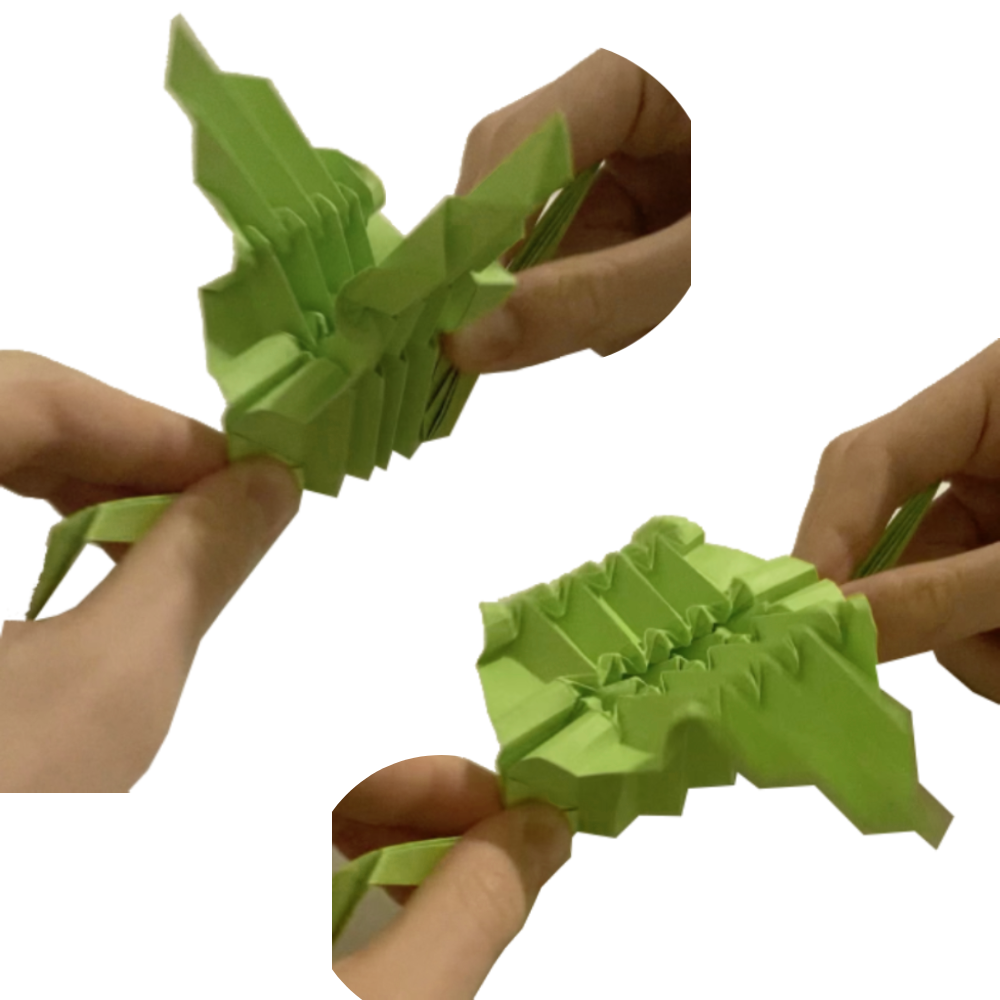}
    }
    \caption{Origami cranes augmented with spring joints to engineer complex motion patterns (a) Simple. (b) Complex.}
    \label{fig:origami}
\end{figure}

Origami as an art form, unlike Engineering, is intentionally very restrictive. The requirement of only being able to fold from a single unbroken sheet of paper means that all of the traditional mechanisms found in engineering cannot apply to action origami models. The spring joint, as described in this paper, can be used as a mechanism to engineer motions within pure origami, and so serves as a replacement for traditional mechanisms like gears ratios when trying to create mechanical advantage in origami. For example, by grafting pleats into a traditional crane, spring joints can be added to create motion along multiple axis for a more realistic flapping pattern, shown in figure \ref{fig:origami}. 

\section{Conclusion}
\label{sec:conclusions}
The spring joint mechanism provides a programmable, amplified alternative to the reverse fold in origami mechanisms. This high velocity ratio is important in compliant mechanisms and origami models in order to make them more responsive. And the program-ability by the mathematical model allows for arbitrary angles to be formed by deploying this mechanism. For these reasons I have shown that it has applications in the engineering of new origami structures.

% reference section
\bibliographystyle{osmebibstyle}
\bibliography{osmerefs}

% author affiliations are appended at end of paper
\theaffiliations

\end{document}